\documentclass[runningheads]{llncs}

\usepackage{multirow}
\usepackage{booktabs}
\usepackage{graphicx}
\usepackage{amsmath}

\usepackage{mathrsfs}
\usepackage{amssymb}

\usepackage{algorithm}
\usepackage{algpseudocode}
\usepackage{marvosym}

\usepackage[T1]{fontenc}

\usepackage[colorlinks=true, linkcolor=blue, citecolor=blue,urlcolor=blue]{hyperref}

\begin{document}
\title{NDRL: Cotton Irrigation and Nitrogen Application with Nested Dual-Agent Reinforcement Learning}
\titlerunning{NDRL: Nested Dual-Agent Reinforcement Learning}


\author{Ruifeng Xu\inst{1,2,3} \and 
Liang He\inst{1,2,3,4}\textsuperscript{(\Letter)}}

\authorrunning{R. Xu et al.}

\institute{
1. School of Computer Science and Technology, Xinjiang University,\\ 
Urumqi 830017, China \and
2. Xinjiang Multimodal Information Technology Engineering Research Center,\\
Xinjiang University, Urumqi 830017, China \and
3. Xinjiang Key Laboratory of Signal Detection and Processing,\\ 
Xinjiang University, Urumqi 830017, China \and
4. Department of Electronic Engineering, Tsinghua University,\\ 
Beijing 100084, China\\
\email{\textsuperscript{\Letter}heliang@mail.tsinghua.edu.cn} \\
\email{xuruifeng@stu.xju.edu.cn}
}

\maketitle             
\begin{abstract}
Effective irrigation and nitrogen fertilization have a significant impact on crop yield. However, existing research faces two limitations: (1) the high complexity of optimizing water-nitrogen combinations during crop growth and poor yield optimization results; and (2) the difficulty in quantifying mild stress signals and the delayed feedback, which results in less precise dynamic regulation of water and nitrogen and lower resource utilization efficiency. To address these issues, we propose a Nested Dual-Agent Reinforcement Learning (NDRL) method. The parent agent in NDRL identifies promising macroscopic irrigation and fertilization actions based on projected cumulative yield benefits, reducing ineffective explorationwhile maintaining alignment between objectives and yield. The child agent's reward function incorporates quantified Water Stress Factor (WSF) and Nitrogen Stress Factor (NSF), and uses a mixed probability distribution to dynamically optimize daily strategies, thereby enhancing both yield and resource efficiency. We used field experiment data from 2023 and 2024 to calibrate and validate the Decision Support System for Agrotechnology Transfer (DSSAT) to simulate real-world conditions and interact with NDRL.  Experimental results demonstrate that, compared to the best baseline, the simulated yield increased by 4.7\% in both 2023 and 2024, the irrigation water productivity increased by 5.6\% and 5.1\% respectively, and the nitrogen partial factor productivity increased by 6.3\% and 1.0\% respectively. Our method advances the development of cotton irrigation and nitrogen fertilization, providing new ideas for addressing the complexity and precision issues in agricultural resource management and for sustainable agricultural development.

\end{abstract}

\section{Introduction}
Cotton is one of the most critical agricultural products globally, and its economic value is significant~\cite{Constable_Bange_2015}. Detailed management of cotton is crucial for increasing yield. Cotton growth is primarily influenced by factors such as irrigation levels, fertilization, climate conditions, crop varieties, and soil types~\cite{Li_2018,Shrivastava_2015}. Recently, the interaction between reinforcement learning and crop models has garnered increasing attention from researchers interested in determining optimal agricultural management strategies~\cite{Cai_2023,Wang_2024}. Chen et al.~\cite{Chen_2024} utilized irrigation dates and average yield as states and employed value iteration strategies to construct a cotton irrigation system. Chen and Cui et al.~\cite{Chen_2021} combined short-term weather forecasts with a Deep Q-Learning reinforcement learning algorithm to determine optimal irrigation strategies. All of the aforementioned algorithms optimize irrigation strategies by controlling a single irrigation amount, without considering the interactive effects between fertilization and irrigation. Tao et al.~\cite{Tao_2023} were the first to use reinforcement learning for simultaneous irrigation and fertilization optimization. However, the action space in their study was limited to a small number of discrete actions, and they used a uniform distribution for action selection during exploration, resulting in significant inefficiency~\cite{Agyeman_2024}. Moreover, all the above - mentioned works focus on trajectory sequences in their reward settings and directly penalize irrigation and fertilization actions. They fail to consider dynamic feedback from crop stress signals, which can lead to unbalanced resource use and potential over-utilization.

Based on the aforementioned research and challenges, we propose a nested dual-agent reinforcement learning with cotton irrigation and fertilization. (NDRL). Additionally, nitrogen (N) is the most critical nutrient input in cotton production and plays a vital role in determining yield and quality~\cite{Tian_2024}. Therefore, this study primarily focuses on optimizing the amount of nitrogen fertilizer. The specific contributions of the algorithm are as follows:

\begin{itemize}
\item To solve the problem of poor yield optimization due to the high complexity of water - nitrogen combination optimization, we use a nested dual - agent method to jointly optimize irrigation and nitrogen fertilizer strategies. The parent agent identifies potentially high - quality macroscopic irrigation and fertilization actions based on projected cumulative yield benefits, reducing ineffective actions. The child agent uses a mixed probability distribution within the daily neighborhood space to further optimize the macroscopic parent actions.

\item To tackle the issue of low resource efficiency, we incorporate DSSAT-quantified water stress factor (WSF) and nitrogen stress factor (NSF) into the child reward function. This allows us to use crop yield as a positive reward and apply stress based penalties, thereby enhancing water-nitrogen productivity.

\item Experimental results demonstrate that, compared to the best baseline, the simulated yield increased by 4.7\% in both 2023 and 2024, the irrigation water productivity increased by 5.6\% and 5.1\% respectively, and the nitrogen partial factor productivity increased by 6.3\% and 1.0\% respectively. 
\end{itemize}

\section{Related Work}

\subsection{Crop Models for RL}
Field experiments are time-consuming and costly. Assessing the impact of climate change and farm practices on crop production using crop models is crucial for agricultural research~\cite{Tobin_2017}. APSIM and DSSAT, the two most widely used crop models, can simulate crop production and estimate it accurately based on climate, genotype, soil, and management factors~\cite{Ara_2021,Holzworth_2014}. Reinforcement learning (RL), which aims to identify optimal strategies through continuous interaction with an environment, can be integrated with crop models~\cite{Mei_2023}. This integration allows for the simulation of production while continuously evaluating management strategies. Such an approach mitigates the drawbacks of real field experiments, validates algorithm advantages, and applies optimal management strategies in practice~\cite{Yang_2022}. 

\begin{figure}[ht]
    \centering
    \includegraphics[width=0.48\linewidth]{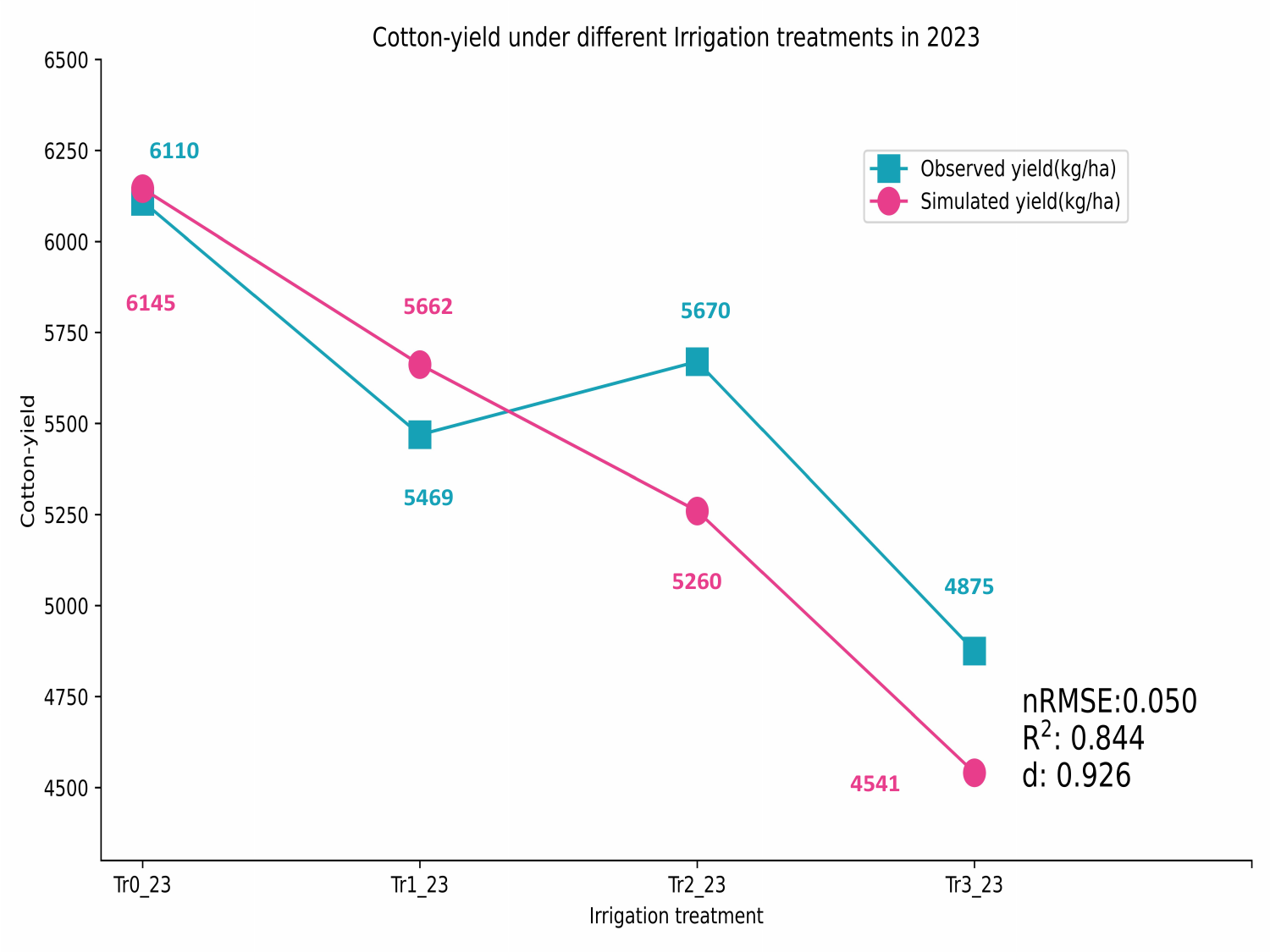}
    \includegraphics[width=0.48\linewidth]{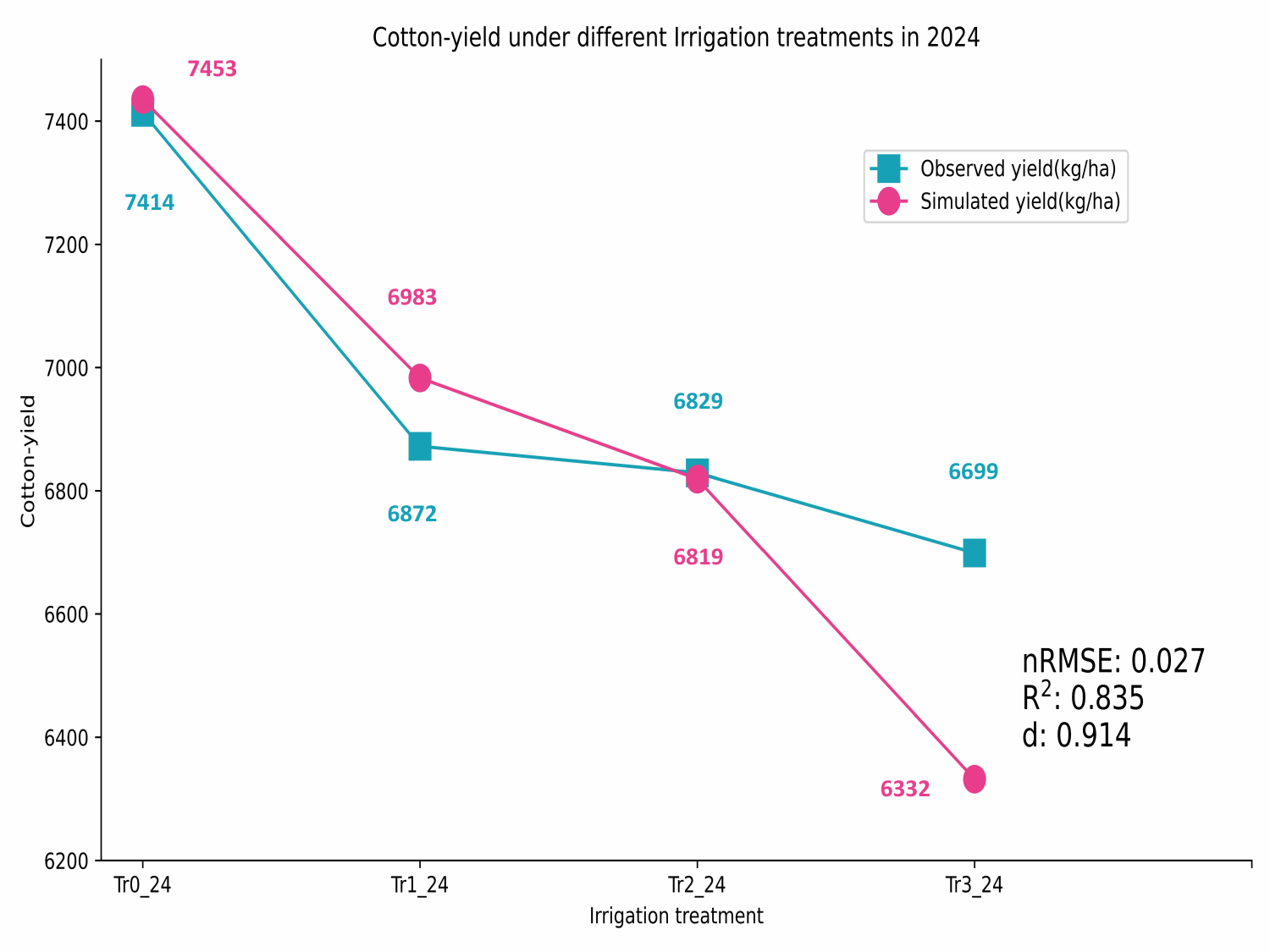}
    \caption{Simulated yield and observation results.}
    \label{fig:Simulated_2023_2024}
\end{figure}

To this end, we have selected DSSAT as our simulation environment and calibrated it using real cotton experimental field data from Huaxing Farm in Changji City, Xinjiang, China (44$^\circ$22$'$N, 87$^\circ$29$'$E), as shown in Table ~\ref{tab:treatment}. As can be seen from Fig~\ref{fig:Simulated_2023_2024}, the normalized root mean square error (nRMSE) values for yields in 2023 and 2024 are 5.0\% and 2.7\%, respectively, both of which are less than 10\%, indicating very good simulation accuracy. The d values are very close to 1, and the R$^2$ values are both greater than 60\%. Additionally, we calculated the nRMSE values between the simulated and actual values for flowering and maturity dates in 2023 and 2024, and the results were all less than 10\%. These findings suggest that the calibrated DSSAT model is suitable for simulating the local environment~\cite{Rugira_2021}.

\begin{table*}[h]
    \centering
    \caption{The table provides a detailed presentation of the irrigation and nitrogen fertilizer application amounts for different time periods in 2023 and 2024}
   
    \label{tab:treatment}
    \resizebox{\textwidth}{!}{
    \begin{tabular}{ccccccccccccccc}
    \toprule
   \multirow{2}{*}{Treatment} & \multicolumn{12}{c}{ Irrigation and Nitrogen Fertilization Dates (2023/Month/Day) } &Total Amount &\multirow{2}{*}{Yield (kg/ha)}\\\cmidrule{2-13}  &04/20&06/09&06/19&06/29&07/09&07/14&07/21&07/28&08/04&08/11&08/18&08/25&(mm,kg/ha)&\\
     \midrule
     \multirow{2}{*}{Tr0\_23}&45&52&37&37&45&37&60&52&52&45&45&30&537&\multirow{2}{*}{6110}\\
     &0&16&25&27&33&41&53&53&0&2&0&0&250&\\  
      
     \multirow{2}{*}{Tr1\_23}&40&47&33&33&41&33&54&47&47&41&41&27&484&\multirow{2}{*}{5469}\\
     &0&16&25&27&33&41&53&53&0&2&0&0&250&\\
    \multirow{2}{*}
    {Tr2\_23}&38&44&31&31&38&31&51&44&44&38&38&26&454&\multirow{2}{*}{5670}\\
     &0&16&25&27&33&41&53&53&0&2&0&0&250&\\
     \multirow{2}{*}
    {Tr3\_23}&36&41&30&30&36&30&48&41&41&36&36&24&429&\multirow{2}{*}{4875}\\
     &0&16&25&27&33&41&53&53&0&2&0&0&250&\\
     \addlinespace
     \midrule
     \multirow{2}{*}{Treatment} & \multicolumn{12}{c}{ Irrigation and Nitrogen Fertilization Dates (2024/Month/Day) } & Total Amount&\multirow{2}{*}{Yield (kg/ha)}\\\cmidrule{2-13}  &05/01&06/07&06/19&06/29&07/09&07/14&07/22&07/30&08/07&08/11&08/21&08/25&(mm,kg/ha)&\\
     \midrule
      \multirow{2}{*}
      {Tr0\_24}&45&52&37&37&45&37&60&52&52&45&45&30&537&\multirow{2}{*}{7414}\\
      &0&15&8&24&45&45&45&53&15&0&0&0&250\\
      \multirow{2}{*}
      {Tr1\_24}&45&52&37&37&45&37&60&52&52&45&45&30&537&\multirow{2}{*}{6872}\\
      &0&13&7&22&41&41&41&47&13&0&0&0&225\\
       \multirow{2}{*}
      {Tr2\_24}&45&46&33&40&47&47&40&40&47&33&27&0&445&\multirow{2}{*}{6829}\\
       &0&15&8&24&45&45&45&53&15&0&0&0&250\\
       \multirow{2}{*}
      {Tr3\_24}&45&46&33&28&46&46&40&40&46&33&27&0&430&\multirow{2}{*}{6699}\\
       &0&15&8&24&45&45&45&53&15&0&0&0&250\\
      \bottomrule
    \end{tabular}
   }
\end{table*}

\subsection{Crop Management with RL}
Regarding crop management as a Markov Decision Process (MDP) and using reinforcement learning to interact with simulators to find optimal strategies has been applied to some extent, but the overall research is still in its early stages~\cite{Sidiropoulos_2024,Agyeman_2025}.Tao et al. applied Deep Q-Networks to jointly optimize irrigation and fertilization\cite{Tao_2023}. However, their use of a global uniform discrete action selection strategy resulted in low exploration efficiency and restricted potential for long-sequence water-fertilizer optimization. Balderas et al.~\cite{Balderas_2025} comprehensively compared DQN and PPO in combined water-nitrogen management, showing the advantage of DQN in such tasks. This study also treats irrigation and fertilization as direct punishment items, ignoring the potential effects of crop stress information.

To address these challenges, inspired by Double Q-learning~\cite{Diddigi_2022} and the multi-agent algorithm minimax-Q~\cite{Vezhnevets_2017}, we propose a dual-agent architecture to achieve efficient exploration and local refinement while optimizing the reward function design, so as to realize more efficient and precise agricultural resource management.

\section{Methodology}
Q-value-based reinforcement learning algorithms, which guide the optimization of water-nitrogen management strategies through state-action value functions, have been demonstrated as effective in prior studies ~\cite{Mnih_2015,Agyeman_2025}. Building upon these foundations, our research introduces innovations in Q-value-based reinforcement learning algorithms and integrates them with the DSSAT (Decision Support System for Agrotechnology Transfer) model to establish a comprehensive model framework. The detailed methodology will be elaborated in the following sections.

\subsection{MDP Problem Formulation for Crop Management}
This study focuses on optimizing irrigation and nitrogen fertilizer application rates to improve crop yield under fixed irrigation-Nitrogen schedules, modeled as a finite Markov Decision Process (MDP) within a hierarchical reinforcement learning framework.

To achieve efficient optimization, we define the state space of the parent agent as a low-dimensional discrete space $ S_p = \{s_{p1}, \dots, s_{pi}\}$, where each parent state $ s_{pt} = (\text{DAYS}, \text{P\_Act\_dis})$. We employ Q-learning for efficient policy updates, with the Q-value update rule given by:
\begin{equation}\label{eq:q-learning}
Q(s_{pt}, a_{pt}) \leftarrow Q(s_{pt}, a_{pt}) + \alpha \left[ R_{pt} + \gamma \max_{a_{p}'\in \mathcal{A}{p}} Q(s_{pt}', a_{pt}') - Q(s_{pt}, a_{pt}) \right]
\end{equation}
The state space of the child agent is defined as a high-dimensional state space $ S_c = \{s_{c1}, \dots, s_{cj}\} $. Each child state $ s_{cz} = (\text{DAY}, \text{WSF}, \text{NSF},\text{LAID} )$ where WSF and NSF record whether stress is present or not; if stress is present, it is 1, otherwise it is 0. We use DQN for adaptive updates, employing a three-layer network architecture. The online network $Q(S_{ct}, a_{ct}; \theta)$ predicts action values in real time, while the target network $Q(S_{ct}, a_{ct}; \theta^-)$ updates parameters with a delay. The loss function is as follows:
\begin{equation}\label{DQN}
    L(\theta) = \mathbb{E}_{(s_{cz},a_{cz},r_{cz},s'_{cz})} \left[ \left( r_{cz} + \gamma \cdot \max_{a'_{cz}} Q(s'_{cz}, a'_{cz}; \theta^-) - Q(S_{cz}, a_{cz}; \theta) \right)^2 \right]
\end{equation}
The parameters of the target network $\theta^-$ are periodically synchronized from the online network.

Given that there are only 12 irrigation and fertilization events throughout the entire growth cycle, we define the macro-trajectory as a policy chain composed of 6 sequential decisions covering the entire growth period of the crop. The micro-trajectory focuses on short-term response optimization with 2 adjacent decisions. The specific parameters of the agent's state space are detailed in Table~\ref{tab: state space}, and all states as well as the yield are obtained from DSSAT.

\begin{table*}[ht]
\centering
\caption{The definition of the state space corresponding to the parent agent and the child agent. YYDDD is a 5-digit compact date code formed by the last two digits of the year followed by the ordinal day of that year. }
\label{tab: state space}
\resizebox{\textwidth}{!}{
\begin{tabular}{lp{3cm}p{8cm}}
\toprule
 & State Variable &  Description \\
\midrule
\multirow{3}{*}{Parent State Space}
& DAYS & The collection of adjacent irrigation and nitrogen fertilization dates (YYDDD) \\
& P\_Act\_dis & The macro action pairs corresponding to the adjacent dates (mm,kg/ha)\\

\\
\multirow{6}{*}{Child State Space}
& DAY & Irrigation and fertilization date (YYDDD)\\
& WSF & The average value of the water stress factor corresponding to the date (0 or 1)\\
& NSF & The average value of the nitrogen stress factor corresponding to the date (0 or 1)\\
& LAID & Leaf area index\\
\bottomrule
\end{tabular}
}
\end{table*}

\subsection{Nested Hierarchical Action Selection}
In the water - nitrogen management task, the action space is the combination of irrigation and nitrogen fertilizer application. Its size increases exponentially compared to that of a single variable. To address this, we propose a nested hierarchical action selection method to explicitly increase the exploration rate. 

During the exploration phase, a model - based strategy is employed to identify potentially good actions for updating the parent agent. Specifically, the DSSAT model predicts the expected cumulative yield $Y(a_p, s_{pt})$ for all parent actions $a_{p} \in {A_p}$. A candidate action set is selected by choosing actions that maximize Y, and the parent action is obtained via a uniform distribution from this set. The parent action is selected as follows:
\begin{equation}\label{parnet_action}
a_{pt} \sim \mathcal{U}\left({ a_p \in {A}_p \mid Y(a_p \mid s_{pt}) \geq \eta \cdot \max_{a} Y(a \mid s_{pt}) } \right)
\end{equation}
Here, $a_{pt}=(I_1,N_1,I_2,N_2)$ where $I$ and $N$ represent irrigation and nitrogen fertilizer application rates, respectively, and the subscripts indicate the order of two consecutive state dates. $\eta$ is set to 0.8.

The child agent refines the macro action \(a_{pt}\) using an \(\epsilon\)-greedy strategy. The child action space \({A}_{cz}\) is defined as 25 equidistant action points discrete from the dynamic asymmetric neighborhood space \([\max(a_{\text{ptz}} - \Delta, (0,0)), \min(a_{\text{ptz}} + \Delta, (I_{pmax},N_{pmax}))]\) centered on the daily macro action \(a_{ptz} = (I_z, N_z) \in a_{pt}\). \(I_{p\text{max}}\) and \(N_{p\text{max}}\) are the maximum irrigation and nitrogen fertilizer application rates selectable by the parent agent.

It's believed that the optimal child action should fluctuate around the central action in the child action space and follow a discrete truncated Gaussian distribution. To this end, the negative exponential terms of the squared Euclidean distance between each candidate action \(a_{cz}=(I_{cz},N_{cz})\) and the central action \(a_{cc}=(I_{cc},N_{cc})\) are calculated to construct a truncated Gaussian probability distribution on the discrete action space. Also, softmax normalization is applied to ensure the sum of all action probabilities equals 1. The distribution is given by:
\begin{equation}
P_\text{Gauss}(a_{cz}) = \frac{\exp\left(-\frac{(I_{cz} - I_{cc})^2 + (N_{cz} - N_{cc})^2}{2\sigma^2}\right)}{\sum_{a_{cz}' \in {A}_{cz}} \exp\left(-\frac{(I_{cz}' - I_{cc})^2 + (N_{cz}' - N_{cc})^2}{2\sigma^2}\right)}
\end{equation}

Where $\sigma={\Delta}/2$. Moreover, to enhance the exploration efficiency of marginal actions, we mix the uniform distribution with the Gaussian distribution in proportion. The formula is as follows:
\begin{equation}
P(a_{cz}) = {(1 - \alpha) \cdot P_{\text{Gaussian}}(a_{cz})} + {\alpha \cdot \frac{1}{|{A}_{cz}|}}
\end{equation}
where \(\alpha = 0.6\), and \(\alpha \cdot \frac{1}{|{A}_{cz}|}\) denotes the uniform distribution. The specific \(\epsilon\)-greedy strategy is as follows:
\begin{equation}
a_{cz} =
\begin{cases}\label{child_action}
\underset{a_{cz} \in {A}_{cz}}{\arg\max} \, Q(s_{cz}, a_{cz}) , & \text{with probability } 1 - \epsilon \\
a \sim P(a_{cz}) , & \text{with probability } \epsilon
\end{cases}
\end{equation}

Through this design, when the irrigation and nitrogen fertilizer application ranges are both 0 - 60 and \(\Delta\) is (20,20), the parent agent's action space \({A}_P\) consists of 256 actions. According to the parent agent's action selection mechanism, the set of potentially good actions during exploration contains approximately 50 actions. The child agent's action space is generated based on the macro action \(a_{ptz}\), and the specific child action \(a_{cz}\) is obtained according to the probability distribution. Compared to simple random action selection during exploration, our method explores high - value areas more effectively and achieves more precise control.

\subsection{Reward Mechanism Design}
We carefully designed the parent reward function. When resources are exceeded, punishment is significantly increased. If the cumulative yield benefit of the child agent after optimization is not higher than that of the parent agent before optimization, it means the child agent's optimization is invalid, and we designed corresponding rewards. In other cases, we use the maximum Q value of the child reward function as the reward to ensure the parent and child agents have the same goal. The formula is as follows:
\begin{equation}
r_{pt} = 
\begin{cases}

0 & \text{if } \sum_{\tau=1}^t I_{{u},\tau} < I_{\text{total}} \land \sum_{\tau=1}^t N_{{u},\tau} < N_{\text{total}} \land H_{\text{c}} \leq H_{\text{p}} \\
-5370 & \text{if } \sum_{\tau=1}^t I_{{u},\tau} \geq I_{\text{total}} \lor \sum_{\tau=1}^t N_{{u},\tau} \geq N_{\text{total}}\\
Q_{\text{cmax}} & Otherwise
\end{cases}
\end{equation}

Here,$\sum_{\tau=1}^t I_{{u},\tau}$ and $\sum_{\tau=1}^t N_{{u},\tau}$ represent the cumulative optimal irrigation and nitrogen fertilizer application from the start to the current state in a complete trajectory sequence. $H_{c}$ and $H_p$ denote the best cumulative yield of the current child agent after optimization and that of the parent agent before optimization, respectively. $Q_{cmax}$represents the maximum Q-value of the child agent.

To support the parent-level decision-making in achieving the goal of "increasing yield and optimizing resource use, and inspired by the work of Chen et al.~\cite{Chen_2024} and Tao et al.~\cite{Tao_2023}, we set the yield as the baseline reward for the child agent. This establishes yield as the primary optimization target. Through coupling the water stress factor (WSF) and nitrogen stress factor (NSF), the child agent is guided to minimize waste when resources are abundant and to increase input when resources are scarce.The specific formula for the child reward is as follows:
\begin{equation}
r_{cz} = \mathrm{HWAM} + \sum_{x \in \{I_{cz}, N_{cz}\}} \left( w_x \cdot \mathbb{I}_{\text{reward},x} - w_x \cdot \mathbb{I}_{\text{penalty},x} \right)
\end{equation}
The formulas for the reward and penalty conditions are:
\begin{equation}
\mathbb{I}{\text{penalty},x} =
\begin{cases}
1 & \text{if } \left( \text{WSF}=0 \land \text{NSF}=0 \land x > \text{avg}(x{\text{total}}) \right) \lor \\ & \quad \left( \text{WSF}>0 \land \text{NSF}>0 \land x < \text{avg}(x_{\text{total}}) \right) \lor \\ & \quad \left( (\text{WSF}=0 \lor \text{NSF}=0) \land x > \text{avg}(x_{\text{total}}) \right) \\
0 & \text{otherwise}
\end{cases}
\end{equation}
\begin{equation}
\mathbb{I}{\text{reward},x} =
\begin{cases}
1 & \text{if } \left( \text{WSF}>0 \land \text{NSF}>0 \land x > \text{avg}(x{\text{total}}) \right) \lor  \\& \quad \left( \text{WSF}>0 \land x > \text{avg}(x_{\text{total}}) \right) \lor  \\ & \quad \left( \text{NSF}>0 \land x > \text{avg}(x_{\text{total}}) \right) \\
0 & \text{otherwise}
\end{cases}
\end{equation}
We set the irrigation and nitrogen fertilizer application of the field control group as I\_total and N\_total, at 537 mm and 250 kg/ha respectively. The values of WSF and NSF range from 0 to 1, where 0 means "no stress" and 1 means "stress".

\begin{algorithm}[ht]
\caption{NDRL for Water-Nitrogen Management}
\label{alg:enhanced_hierarchical}
\begin{algorithmic}[1]
\Require Parent Q-table $Q_p$, Child DQN parameters $\theta$, Target network parameters $\theta^-$, Replay buffer $\mathcal{B}$,\\ 
\hspace*{2em}Exploration parameters $\eta=0.8$, $\alpha = 0.6$.
\Ensure Optimized policies $\pi_p^*$, $\pi_{c}^*$

\Statex \textbf{Initialization:}
\State Initialize parent state $s_{p0} \gets (\text{DAYS}, \text{P\_Act\_dis})$

\While{not converged}
    \For{each macro-cycle $k \in \{1,\dots,6\}$}
        \State \textsc{Parent Action Selection:}
        \State Generate candidate actions: using $\epsilon$-greedy policy to select action $a_{pt}$ during exploration follow the Eq\eqref{parnet_action}
        
        \State \textsc{Child Action Optimization:}
        \For{each micro-step $t \in \{1,2\}$}
            \State Construct child state $s_{cz} \gets (\text{DAY}, \text{WSF}, \text{NSF}, \text{LAID})$
            \State Select child action: follow the Eq\eqref{child_action}
           
            \State Execute action, observe $(r_{cz}, s'_{cz})$
            \State Store transition $(s_{cz}, a_{cz}, r_{cz}, s'_{cz})$ in $\mathcal{B}$
        \EndFor

        \State \textsc{Parameter Updates:}
        \If{$|\mathcal{B}| \geq \text{BatchSize}$}
            \State Sample batch $\{(s^{(i)}, a^{(i)}, r^{(i)}, s'^{(i)})\}_{i=1}^N \sim \mathcal{B}$
            \State Update Q-network following the Eq \eqref{DQN}
        \EndIf
        \State Update parent reward from optimal child policy with associated $Q_{cmax}$
        \State \textsc{Parent Q-Update:}
        \State Update Q-value following the Eq \eqref{eq:q-learning}
    \EndFor
\EndWhile
\end{algorithmic}
\end{algorithm}

\begin{figure*}[ht]
    \centering
    \includegraphics[width=0.9\linewidth]{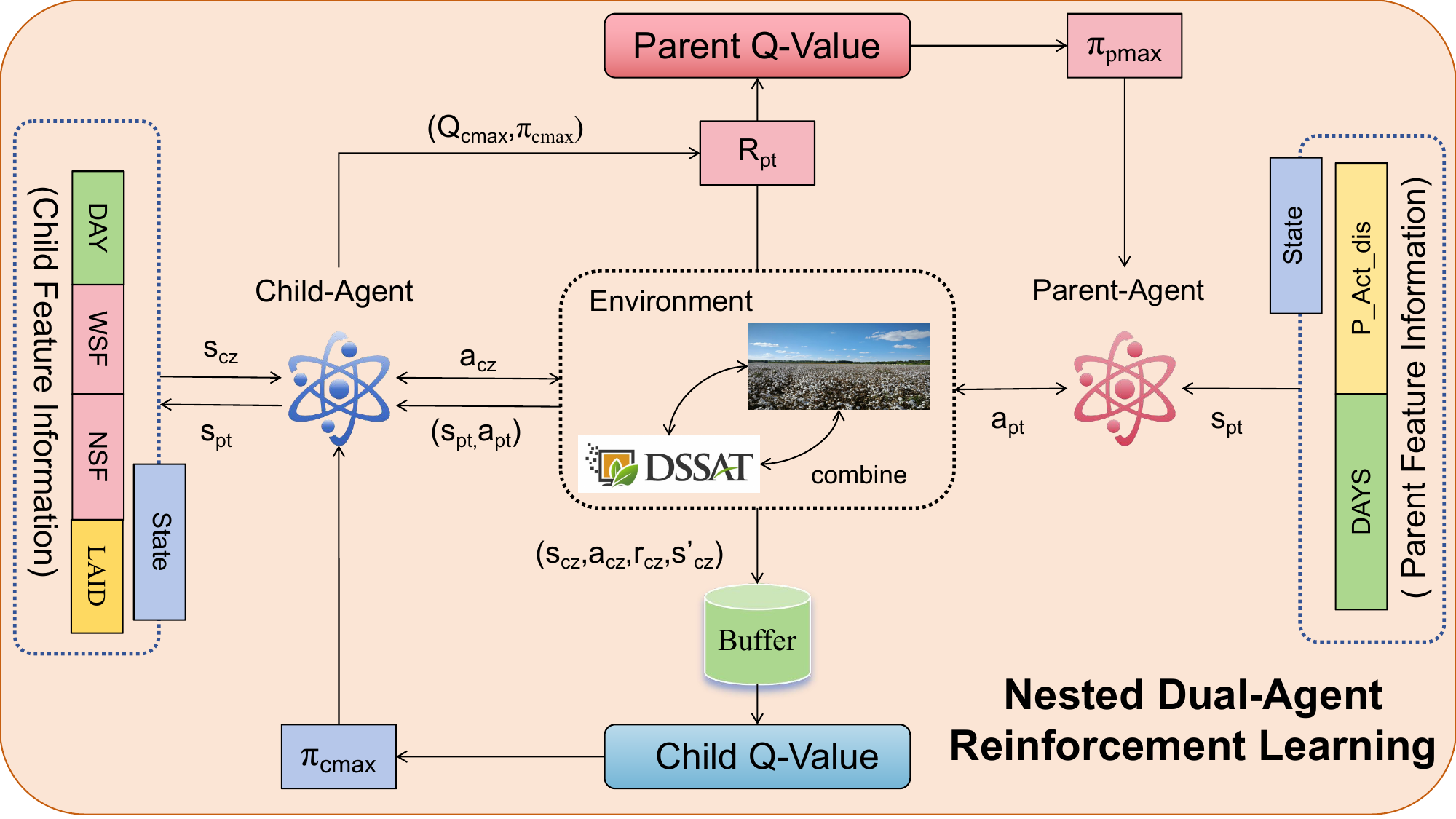}
    \caption{This is the framework of a nested dual-agent reinforcement learning algorithm, which is divided into two parts. The left side represents the optimization process of the child agent, while the right side represents the optimization process of the parent agent. Subscripts beginning with $p$ represent abstractions related to the parent agent, while those beginning with $c$ represent abstractions related to the child agent.} 
    \label{fig:NDRL}
\end{figure*}

\subsection{NDRL Framework}
As shown in Fig~\ref{fig:NDRL} and Algorithm~\ref{alg:enhanced_hierarchical}, the iterative process is as follows: 

The parent agent selects a potentially better macro-action $a_{pt}$ based on the current parent state $s_{pt}$ using an $\epsilon$-greedy strategy and passes it to the child agent. The child agent forms the child state $s_{cz}$ based on the parent action and parent state $s_{pt}$. The child agent continuously selects child actions $a_{cz}$ for interaction. The tuples $(s_{cz}, a_{cz}, r_{cz}, s'_{cz})$ are stored in the buffer for updating the Q-network. The parent reward and state are updated using the action $C\_{\text{Act\_dis}}$ corresponding to the optimal child policy and the associated child Q-value $Q_{cmax}$. The child agent's state is then reset in preparation for the next parent state update.

\section{Experiments}
\subsection{Evaluation Metrics and Baseline}
The experiments in this study were conducted in a calibrated simulation environment. 

\textbf{Baseline.}Our algorithm was compared with real field experiments and the DQN~\cite{Mnih_2015} algorithm. 

\textbf{Evaluation metrics.} The evaluation metrics include both agronomic and algorithm performance indicators. The agronomic indicators are Irrigation Water Productivity (IWP) ~\cite{Baird_2024} and Nitrogen Partial Factor Productivity (NPFP) ~\cite{Lin_2024}, with the specific formulas as follows:
\begin{equation}
\text{IWP} = \frac{\text{Cotton Yield}}{\text{Irrigation Water Volume}}  (\text{kg/m}^3)
\end{equation}
\begin{equation}
\text{NPFP} = \frac{\text{Cotton Yield}}{\text{Nitrogen Fertilizer Application}} (\text{kg/kg})
\end{equation}

\subsection{Comparison of Yield under Different Management Strategies}

This study evaluated the NDRL algorithm, DQN, and real field data using agronomic performance indicators across 2023 and 2024.As shown in Table~\ref{tab:Management}, the following can be observed:

\begin{table*}[ht]
\centering
\caption{Comparison of NDRL with Traditional Reinforcement Learning Algorithms and Real Field Data }
\label{tab:Management}
\resizebox{\textwidth}{!}{ 
\begin{tabular}{@{}llcccccc@{}}
\toprule
\textbf{Category} & \textbf{Year} & \textbf{Methods} & \textbf{Irrigation (mm)} & \textbf{Fertilizer\_N (kg/ha)} & \textbf{Yields (kg/ha)} & \textbf{IWP (kg/m$^3$)} & \textbf{NPFP (kg/kg)} \\

\midrule

\multirow{6}{*}{\textbf{Water Management}} 
 & \multirow{3}{*}{2023} 
 & Field Data & 537 & 250 & 6110 & 1.13 & 24.44 \\
 & & DQN & \textbf{511} & -- & 6255 & 1.22 & -- \\
 & & NDRL  & 512 & -- & \textbf{6272} & \textbf{1.22} & -- \\
\addlinespace
 & \multirow{3}{*}{2024}
 & Field Data & 537 & 250 & 7414 & 1.38 & 29.65 \\
 & & DQN & 509 & -- & 7395 & 1.45 & -- \\
 & & NDRL  & \textbf{488} & -- & \textbf{7443} & \textbf{1.52} & -- \\

\midrule

\multirow{6}{*}{\textbf{Nitrogen Management}} 
 & \multirow{3}{*}{2023}
 & Field Data & 537 & 250 & 6110 & 1.13 & 24.44 \\
 & & DQN & -- & 239 & 6370 & -- & 26.65 \\
 & & NDRL & -- & \textbf{239} & \textbf{6474} & -- & \textbf{27.08} \\
\addlinespace
 & \multirow{3}{*}{2024}
 & Field Data & 537 & \textbf{250} & 7414 & 1.38 & 29.65 \\
 & & DQN & -- & 255 & 7484 & -- & 29.34 \\
 & & NDRL & -- & 251 & \textbf{7562} & -- & \textbf{30.12} \\

\midrule

\multirow{6}{*}{\textbf{Water-Nitrogen Management}} 
 & \multirow{3}{*}{2023}
 & Field Data & 537 & 250 & 6110 & 1.13 & 24.44 \\
 & & DQN & 504 & 252 & 6341 & 1.25 & 25.16 \\
 & & NDRL & \textbf{501} & \textbf{248} & \textbf{6641} & \textbf{1.32} & \textbf{26.77} \\
\addlinespace
 & \multirow{3}{*}{2024}
 & Field Data & 537 & \textbf{250} & 7414 & 1.38 & {29.65} \\
 & & DQN & 504 & 300 & 7802 & 1.54 & 26.00 \\
 & & NDRL & \textbf{488} & 264 & \textbf{7913} & \textbf{1.62} & \textbf{29.97} \\

\bottomrule
\end{tabular}
}
\end{table*}

(1) NDRL outperformed all in cross-year, -algorithm, and -task comparisons. Under joint water - nitrogen optimization, it boosted simulated yields by 4.7\% in both years versus the next - best baseline. IWP rose by 5.6\% and 5.1\%, and FNFP by 6.3\% and 1.0\% in 2023 and 2024, respectively.

(2) In single - variable optimization, NDRL and DQN perform similarly. But in joint optimization, NDRL significantly outperforms DQN. DQN uses global uniform action discretization, covering all inefficient combinations and causing sub - optimal strategies. NDRL's parent agent screens for potentially efficient actions, and the child agent conducts local fine - tuned explorations. This avoids too many ineffective explorations and improves yield and resource efficiency.

(3) When comparing across years, it can be seen that the yield in 2024 was generally higher than that in 2023. This was mainly due to the differences in real-world environmental factors between 2023 and 2024. Under the same irrigation conditions, these differences caused the DSSAT simulation results to vary.

\begin{figure}[ht]
    \centering
    \includegraphics[width=0.9\linewidth]{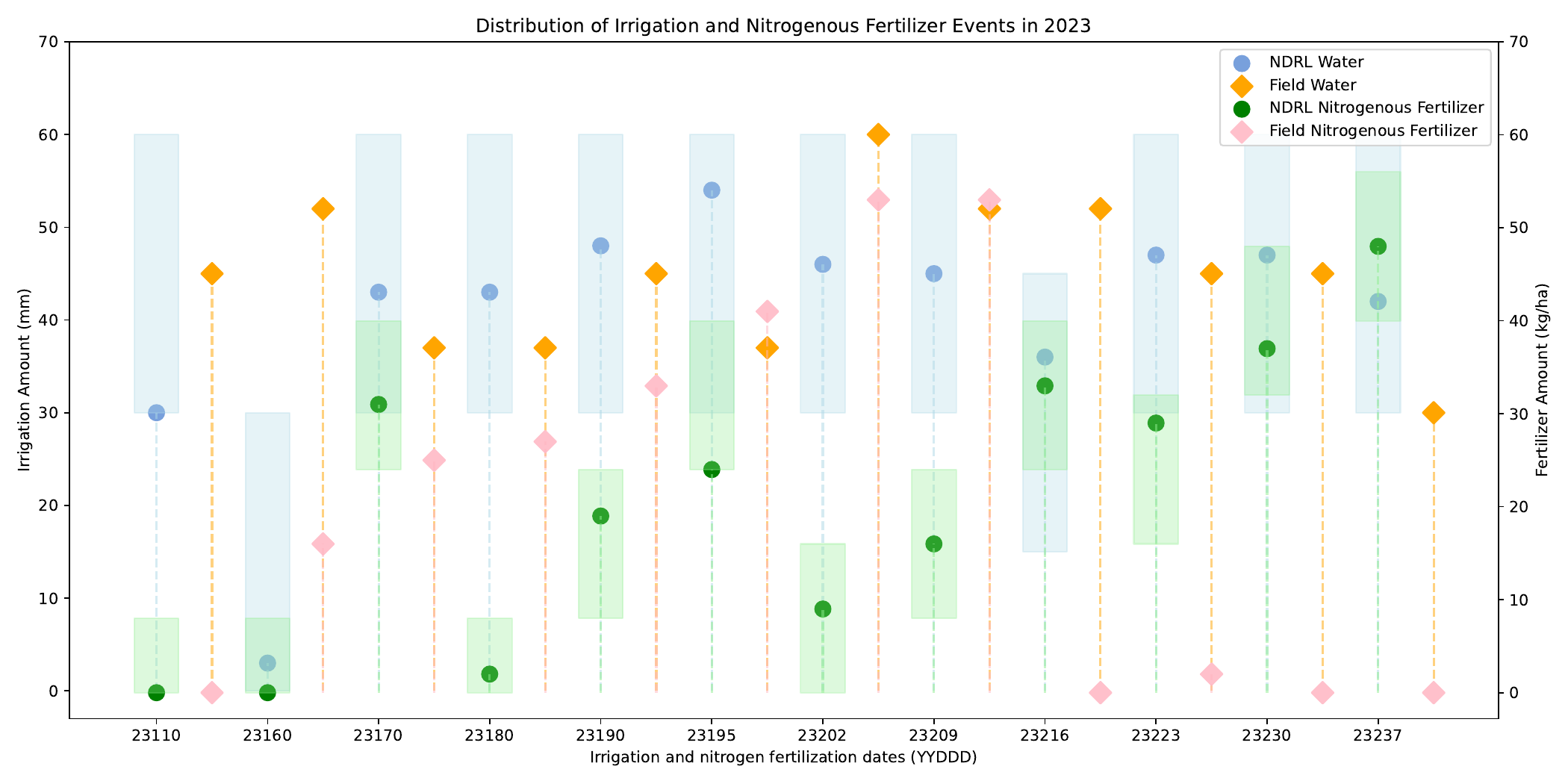}
    \includegraphics[width=0.9\linewidth]{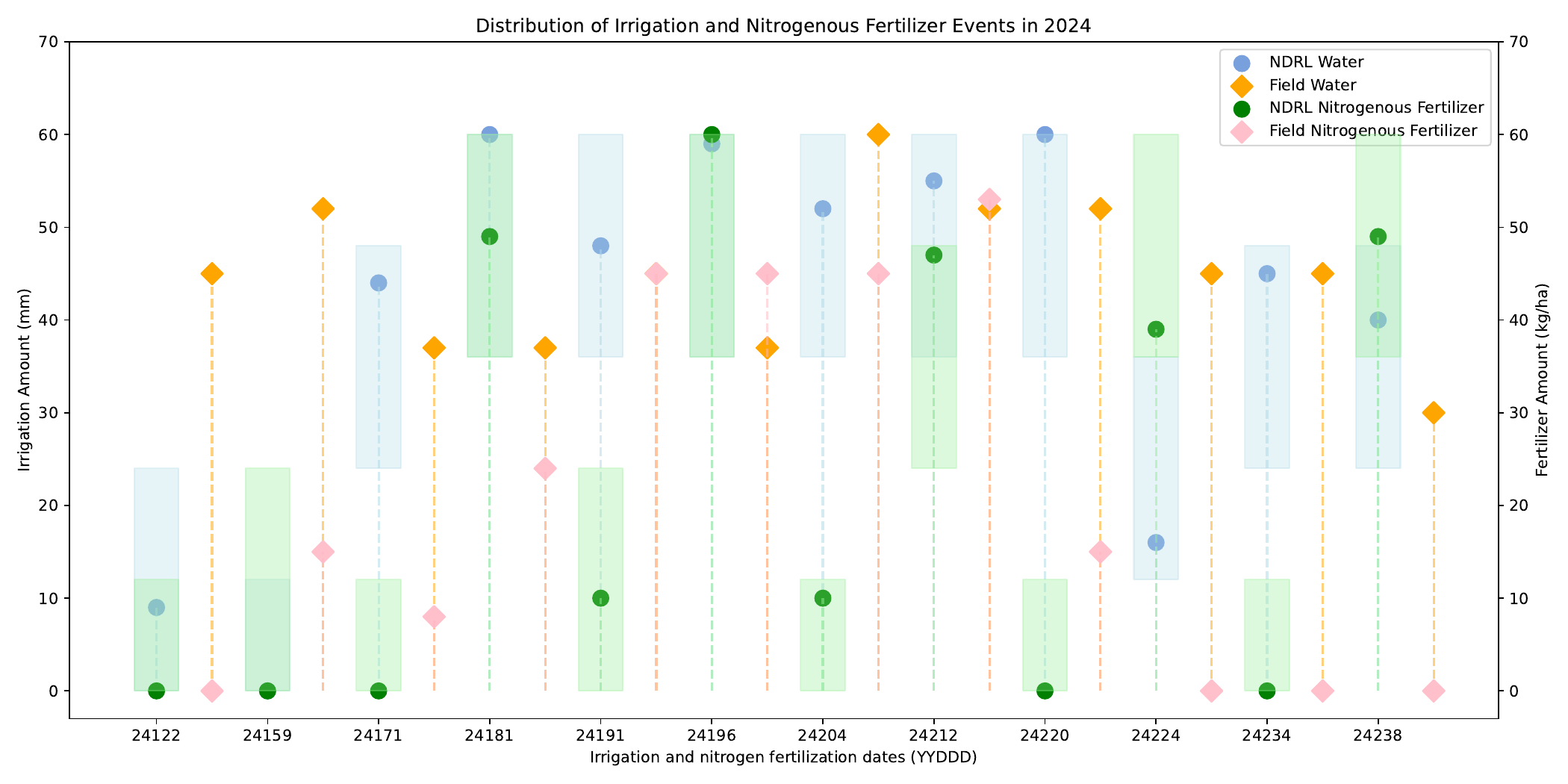}
    \caption{The distribution of irrigation and nitrogen fertilizer application events over two consecutive years reflects the variability and efficiency of the NDRL algorithm in these practices. The x-axis denotes specific dates of irrigation and nitrogen application, the y-axis represents irrigation depth and nitrogen fertilizer amount, and shaded areas indicate the range corresponding to the child agent.}
    \label{fig:Distribution_events}
\end{figure}

\subsection{Distribution of Irrigation and Nitrogen Application Events}
The specific distribution of irrigation and nitrogen fertilizer application times for NDRL under the water-nitrogen integrated task, compared to the field experiment results, is shown in Fig~\ref{fig:Distribution_events}. We found that:

(1) Between planting and the fifth nitrogen fertilizer application during irrigation (seedling stage and boll stage),the amount of nitrogen fertilizer applied increased over time,consistent with the field experiment, and due to the relatively dry seedling stage in 2023, we carried out extensive irrigation as in the field experiment. 

(2) However, during the boll stage, our irrigation and nitrogen fertilizer application slightly differed from the field experiment, which might be the result of optimization in our simulated environment due to the field control experiment.

\subsection{Reinforcement Learning Analysis}
\begin{figure}[ht]
    \centering
    \includegraphics[width=0.48\linewidth]{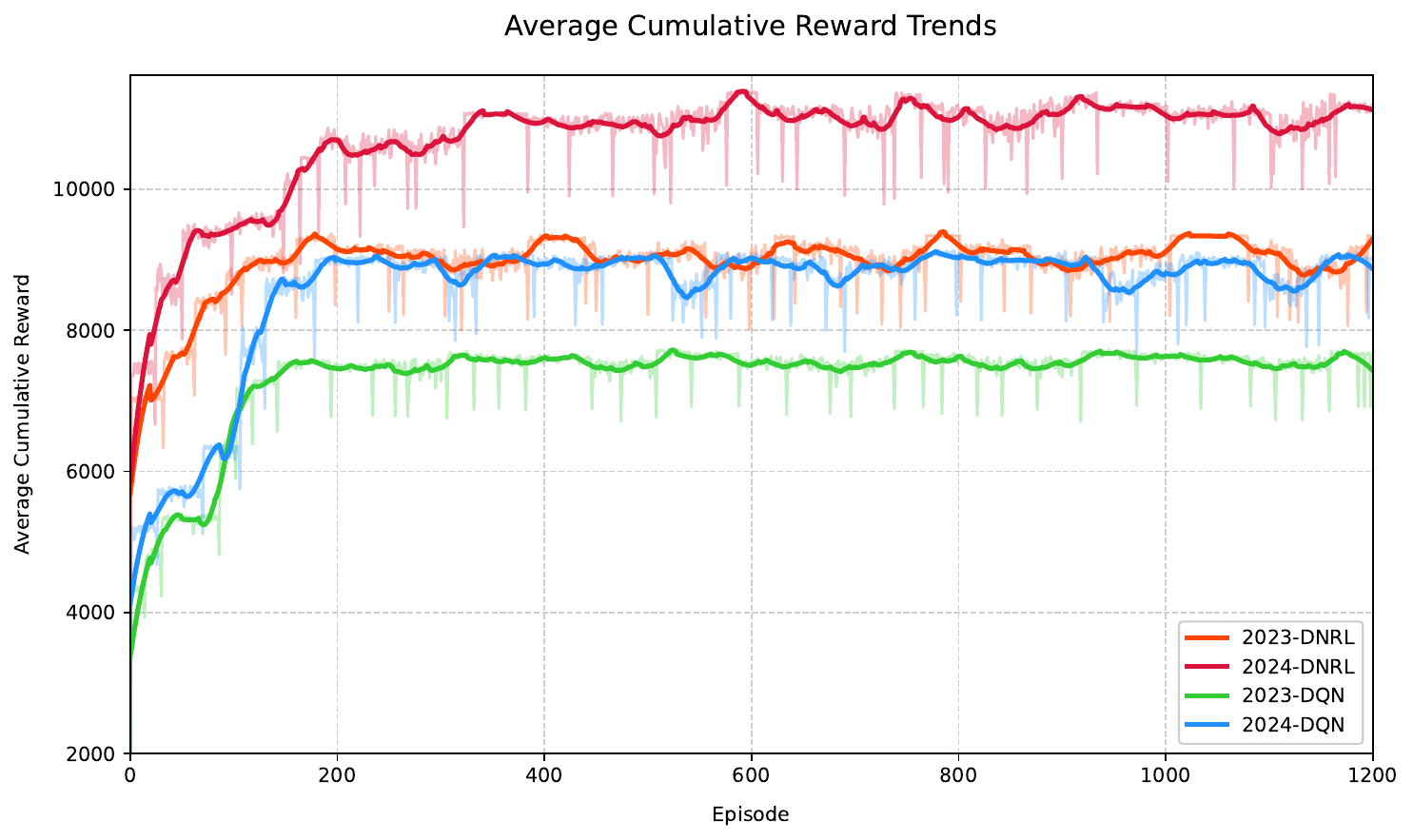}
    \includegraphics[width=0.48\linewidth]{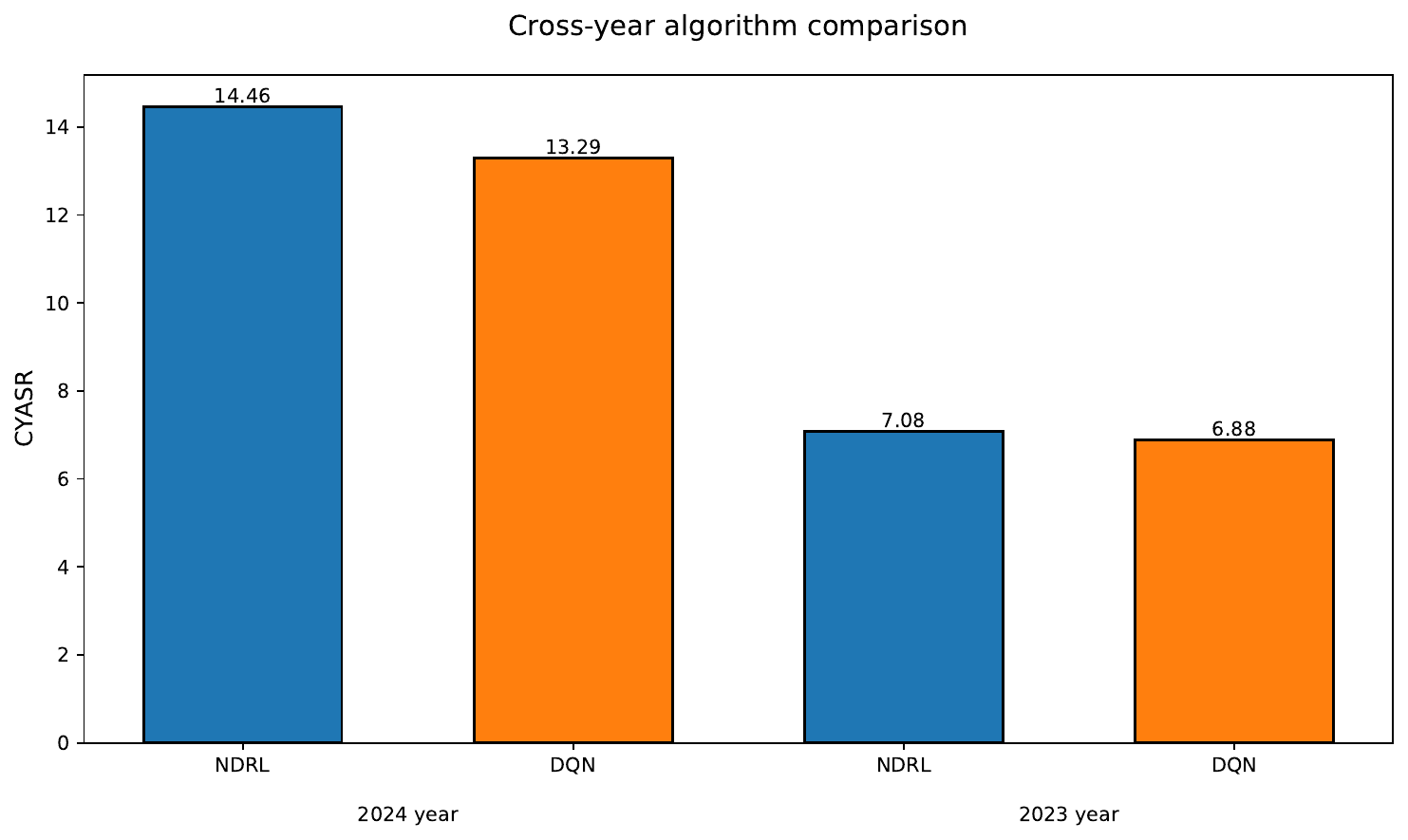}
    \caption{The two graphs are the average cumulative reward trends for 2023 and 2024, and the CYASR comparison between these two years. In the average cumulative reward trends, the x-axis shows training episodes and the y-axis shows average cumulative rewards.}
    \label{fig:Yield Convergence Rate}
\end{figure}

To thoroughly assess the NDRL algorithm's agronomic applicability, we adapted the evaluation metrics from Kaloev and Krastev ~\cite{Kaloev_2021} and introduced the "Cumulative Yield-to-Action Step Ratio (CYASR)". CYASR, the ratio of cumulative yield to action steps, measures decision-making cost per unit. We also analyzed the "average cumulative reward trend", a concept from traditional algorithms.
From Fig ~\ref{fig:Yield Convergence Rate}, we observe that:

(1) NDRL has significantly higher average cumulative rewards than DQN, showing its ability to select higher - reward yield - optimization actions and highlighting the parent agent's action selection advantage. Unlike DQN's random action selection during exploration, NDRL's parent’s action selection mechanism enables more interactions in high - value areas for better rewards. However, due to its nested and hierarchical optimization requiring extensive interactions, NDRL's CYASR doesn't significantly outperform DQN within the same training - episode count.

(2) Cross-year analysis shows lower average rewards and CYASR in 2023 than in 2024, highlighting the environmental sensitivity of agricultural reinforcement learning systems. Environmental heterogeneity and cost constraints in field experiments further show the need for algorithm - driven precision management.

\section{Conclusions}
To address the challenges of high complexity in optimizing irrigation and fertilization combinations during the crop growth cycle, poor yield optimization outcomes, difficulties in quantifying mild stress signals, and delayed feedback (which affects the precision of dynamic water-nitrogen regulation and leads to low resource utilization efficiency), this paper proposes a Nested Dual-Agent Reinforcement Learning Algorithm (NDRL). Combined with DSSAT, this algorithm constitutes an efficient water-nitrogen management approach. Within this algorithm, the parent agent makes macroscopic decisions on irrigation and fertilization for the next two days based on expected cumulative yield benefits, while the child agent dynamically adjusts daily strategies according to the water stress factor (WSF) and nitrogen stress factor (NSF), as well as a mixed probability distribution.

Experimental results show that our method outperforms traditional reinforcement learning-based irrigation and fertilization strategies, and can explore more efficient working modes within the same number of training episodes. Compared to the second-best baseline, the simulated yield increased by 4.7\% in both 2023 and 2024; irrigation water productivity increased by 5.6\% and 5.1\%, respectively; and nitrogen partial factor productivity increased by 6.3\% and 1.0\%, respectively. In future research, we will further expand the dimensionality of the state space and optimize management measures throughout the entire crop growth cycle to promote a shift in agricultural management towards a resource-saving and precise regulation model, thereby supporting sustainable agricultural development.

\section*{Declaration of competing interest}
The authors declare that they have no known competing financial interests or personal relationships that could have appeared to influence the work reported in this paper.

\section*{Acknowledgments} 
This study was supported by the National Science and Technology Major Project (Grant No. 2022ZD0115801).

\end{document}